\PassOptionsToPackage{numbers,sort&compress}{natbib}

\documentclass{article}

\usepackage[final]{neurips_2024}

\usepackage[utf8]{inputenc} 
\usepackage[T1]{fontenc}    
\usepackage{hyperref}       
\usepackage{url}            
\usepackage{booktabs}       
\usepackage{amsfonts}       
\usepackage{nicefrac}       
\usepackage{microtype}      
\usepackage{xcolor} 
\usepackage{caption}

\usepackage[english]{babel}
\usepackage{float}
\usepackage{latexsym}
\usepackage{amssymb}
\usepackage{amsmath}
\usepackage{amsthm}

\usepackage{enumitem}
\usepackage{graphicx}
\usepackage{color}
\newcommand{\BibTeX}{B\kern-.05em{\sc i\kern-.025em b}\kern-.08em\TeX}

\title{AI, Pluralism, and (Social) Compensation}

\author{Nandhini Swaminathan\\
Computer Science and Engineering\\
University of California, San Diego\\
La Jolla, San Diego\\
\texttt{nswaminathan@ucsd.edu}
\AND
David Danks\\
Data Science, Philosophy, and Policy\\
University of California, San Diego\\
La Jolla, San Diego\\
\texttt{ddanks@ucsd.edu}}

\begin{document}

\maketitle
\author{%
  Nandhini Swaminathan\thanks{\href{https://orcid.org/0000-0002-9835-6704}{https://orcid.org/0000-0002-9835-6704}} \\
  \textit{Computer Science and Engineering} \\
  \textit{University of California, San Diego}\\
  La Jolla, San Diego \\
  \and
  David Danks\thanks{\href{https://orcid.org/0000-0003-4541-5966}{https://orcid.org/0000-0003-4541-5966}} \\
  \textit{Data Science, Philosophy, and Policy} \\
  \textit{University of California, San Diego}\\
  La Jolla, San Diego
}

\begin{abstract}
One strategy in response to pluralistic values in a user population is to personalize an AI system: if the AI can adapt to the specific values of each individual, then we can potentially avoid many of the challenges of pluralism. Unfortunately, this approach creates a significant ethical issue: if there is an external measure of success for the human-AI team, then the adaptive AI system may develop strategies (sometimes deceptive) to compensate for its human teammate. This phenomenon can be viewed as a form of ``social compensation,'' where the AI makes decisions based not on predefined goals but on its human partner's deficiencies in relation to the team's performance objectives. We provide a practical ethical analysis of the conditions in which such compensation may nonetheless be justifiable.

\end{abstract}

\section{Introduction}
Value pluralism, and more specifically value conflict, poses a significant challenge for AI designers and developers, as we cannot create a single system that fully supports each individual's values \cite{petersen2021ethical}. Said differently, value pluralism is inconsistent with AI monism. At a high level, there are essentially two different strategies in response. First, we could try to convert value pluralism into value monism (from the perspective of the AI) by integrating the different individual values into a single value function. Different techniques have been tried for this strategy, such as the creation of a social preference function~\cite{noothigattu2018voting, chevaleyre2008preference}, or training the reward model with human feedback from diverse populations~\cite{rafailov2024direct}. Second, we could try to convert AI monism into AI pluralism by creating distinct, personalized systems for each individual \cite{kucirkova2017developing,zhu2014mining,brusilovsky2007user, collins2018rethinking}. This approach is thought to eliminate the value pluralism challenge (at least, in theory\footnote{This approach assumes that each individual has a stable, precise set of values to which the AI can adapt. In practice, people's values can vary substantially over time and context.}), though at the cost of technical difficulties (e.g., insufficient data about the individual) and social complications (e.g., different individuals seeing different outputs for the same input). 

In this paper, we argue that this second approach---personalized AI systems---creates a significant novel ethical challenge whenever there is an external measure of success for the human-AI team. Specifically, we show that an adaptive AI will often learn to compensate (in its behavior) for the shortcomings of the human with which it interacts, thereby leading to potentially deceptive behavior by the AI system (Section 2). We then present a conceptual analysis of the conditions in which such deception is ethically justifiable (Section 3), as well as practical challenges created when an AI personalizes in this way (Section 4). One might have hoped that personalization would provide a straightforward way to sidestep the ethical challenges of value pluralism for AI systems. Unfortunately, it can also create novel ethical challenges.

\section{Inevitability of Compensatory Strategies}
Consider a simple case of value pluralism: there is some general goal $G$ that the human-AI team is attempting to achieve, but the human has additional (or different) values such that the human's goal is actually $H$. For example, imagine a doctor and an AI-driven clinical decision support system \cite{comito2022ai}. The general goal of this human-AI system might be providing accurate diagnoses of patients, but the human additionally wants to have the opportunity to perform novel and interesting surgeries (or, in a more ethically questionable scenario, prioritizing the health of patients of one sex over the other). 

One standard way for the AI to adapt is through reinforcement learning (RL).\footnote{In this specific case, we focus on a simple RL case, but the general conclusion about compensation holds for a much wider class of adaptation or personalization methods. For example, compensation is provably the optimal strategy in a signaling game; see Appendix A.} Markov Decision Process (MDP) methods iteratively refine the agent's policy based on environmental feedback, resulting in policies that maximize the expected cumulative reward, mathematically expressed as: 
\[ J(\pi) = \mathbb{E}_\pi \left[ \sum_{t=0}^\infty R(s_t, a_t) \right] \]
where \( Q^*(s, a) \) is the optimal action-value function and  \( \pi^* \) is the learned policy 
\[ \pi^*(s) = \arg\max_{a \in A} Q^*(s, a) \]

In standard RL implementations, the reward function $R$ is grounded in environmental feedback (i.e., success or failure); in our present example, $R$ would thus be based on goal $G$ (in our case, dependent on whether the human-AI team accurately diagnose the patient). Let $\pi^*(\langle s_E, s_O \rangle, a_t )$ denote the optimal AI policy if the human agent behaves optimally $s_O$ (relative to $R$, and hence $G$).\footnote{Assume $\langle s_E, s_O \rangle_t$ does not depend on $a_{t-1}$ (e.g., the MDP sees a sequence of independently drawn cases).} In this case, however, the human will \emph{not} behave optimally, as they have a different overall goal. Let $s_O'$ denote the actual human behavior; that is, $R(\langle s_E, s_O' \rangle_t, a_t) < R(\langle s_E, s_O \rangle_t, a_t)$. If $R(\langle s_E, s_O' \rangle_t, a_t) < R(\langle s_E, s_O' \rangle_t, a_t')$ for some alternative $a_t'$, then the optimal policy $\sigma^* \neq \pi^*$.

That is, we can directly prove that $\pi^*$ is not optimal given $s_O'$. Instead, the policy $\sigma^*$ that selects the optimal action given $s_O'$ (perhaps $a_t'$, but not necessarily) is necessarily different from $\pi^*$. If other agents perform suboptimally---in this case, humans aiming for their personal goal rather than the external goal---then the AI will learn to do as best as it can, even if that requires acting differently than it should given an optimal teammate. 

In these situations, the AI agent will learn to \emph{compensate} for the human's additional or alternative values. For example, if the human doctor wants to do interesting surgeries, then she would presumably judge the patient to be sicker (or more in need of surgery) than they actually are. An RL-based diagnostic AI would learn to compensate by outputting that such patients are less sick than they actually are, since it would have learned that the human doctor will adjust upwards. One could use an RL method that can change its goal or reward function by learning from its human teammate~\cite{hadfield2016cooperative, sadigh2017active}. That is, the RL agent could be capable of changing $R$ in response to user feedback; in a perfect world (from the perspective of personalization), the RL agent would eventually shift from $R_G$ (i.e., the reward function for goal $G$) to $R_H$. If the RL agent's reward function were $R_H$, then the AI would no longer need to compensate. However, inference to the human's goals or values is almost always partial and noisy; there is little reason to expect that the AI agent's will reliably or systematically learn $R_H$, as opposed to some similar-but-different $R$ (e.g., perhaps one that is smoother than $R_H$) \cite{amodei2016faulty,amodei2016concrete}. This approach can decrease the amount of AI compensation, but is unlikely to eliminate it.

The inevitability of compensation by adaptive AI systems is supported by a variety of research demonstrating how AI systems can compensate in task-oriented environments \cite{ castelfranchi2000artificial, park2023ai, lewis2017deal} . For instance,  Christiano et al. \cite{christiano2017deep} observed that an AI system supposedly trained to grasp a ball instead learned to create an illusion of grasping by hovering its hand in front of the ball to satisfy the human reviewer. Or consider CICERO, an AI system that outperforms human experts in the strategic game Diplomacy, whose game transcripts showed the AI system engaging in premeditated deception and lying to its allies for victory~\cite{meta2022human}. 



 \section{Ethics of Compensatory Algorithm}
The phenomenon of AI compensation bears a striking resemblance to human behavioral patterns in team dynamics mirroring the concept of \emph{social compensation}, where one team member takes on additional work due to a belief that others will fail to fulfill their responsibilities~\cite{williams1991social, karau1997effects}. In human-AI teams, this dynamic manifests when the AI agent perceives its human teammate as "behaving non-cooperatively" or "underperforming" and adjusts its output to compensate for these human biases. From a descriptive standpoint, compensatory behavior by an AI should not be surprising, as it represents another instance of a teammate modifying their behavior in response to the perceived shortcomings or biases of another.

From a normative ethical perspective, however, the situation is significantly more complicated. Compensatory behavior arguably impacts the human's autonomy, understood as the right and ability to make one's own decisions and life-plans~\cite{darwall2006value}. If the AI system engages in compensation, perhaps rising to the level of deception, then it is interfering with the human's ability to pursue their values and goals. Of course, this compensation is arguably beneficial as it mitigates the impact of biases and improves overall team efficiency, but it achieves those ends by (indirectly) manipulating the human. Moreover, efforts to teach or train the human so that compensation is not necessary would infringe upon their autonomy in a different way, as such efforts would aim to change their values to be more consistent with others (i.e., it would aim to homogenize the users) \cite{berlin1969four}.

Violations of autonomy are not necessarily unethical, for example, restrictions on autonomy are often justified for convicted criminals \cite{sep-original-position}. Many permissible autonomy infringements involve cases of paternalism~\cite{kleinig1983paternalism}, where one individual $P$ constrains the options of another $C$ for $C$'s benefit but contrary to $C$'s stated preferences. There is significant literature on the conditions in which paternalism is ethically justifiable (e.g.,~\cite{abramson1989autonomy}). Compensatory AI systems might sometimes be paternalistic (e.g., if the human has self-harmful values that the AI should not learn). However, other cases of compensation are not paternalistic in the standard sense of the term, as they involve benefits for a third party \footnote{For a more thorough discussion, see Appendix A2}. For example, compensation for the surgery-preferring doctor is not for the doctor's benefit, but rather for the patient's. In these cases, the AI must determine whether to prioritize the patient's or doctor's values in its personalization, but analyses of paternalism do not apply to those cases. We thus need a more general ethical analysis of the permissibility of compensatory algorithms.

We propose that it is ethically permissible for an algorithm designer/developer to introduce compensatory mechanisms into the algorithm when the following conditions hold:\footnote{We contend only that these are sufficient conditions; there might be other contexts in which compensatory algorithms are ethically permissible.}
\begin{enumerate}
    \item There exists good evidence that the human's specific values have a negative impact on individuals $I$ (i.e., $G$ has value for $I$ that is lost if we pursue $H$)
    \item There exists a justified belief that a reasonable $I$ would consent to the compensation if made aware of it in advance
    \item $G$ has significantly more moral weight than $H$, and could realistically be achieved
    \item Minimal compensation is used commensurate with achieving $G$ 
    \item The AI system actively minimizes the negative effects of the compensatory act
\end{enumerate}
Importantly, $I$ might be the human in the team (as in a case of paternalism), but could also be a third party. The question of $I$'s (hypothetical) consent is a very difficult one, particularly since explicitly asking for consent could serve to undo the benefits of compensation. Methods adapted from social casework~\cite{abramson1989autonomy, mixson1995chapter} could potentially be useful here, as that domain often requires obtaining consent when individuals have internally conflicting values. We also emphasize (point 3) that compensation depends on the external measure of success being more important than the personal goal. And of course, one ought not deceive (or compensate) if there is some other way to reach the moral goal, though in many domains, alternative mechanisms to reduce biases have proven to be ineffective~\cite{hoberman2007medical, keil2007reporting, taylor2022organisational}.

 \section{Practical Issues \& Longer-term Challenges}
Compensatory AI systems will naturally arise when using adaptive or personalized AI systems, if the human's goals and values deviate from some external goal or measure of success. The previous section focused on the ethical challenge of compensation (as potentially undermining autonomy); here we consider three more practical worries, each with some normative implications.

\textbf{Compensation undermining trust}. We often aim for trustworthy AI systems, though there is significant disagreement about exactly what that means. In human-human interactions, social compensation is inversely related with trust~\cite{karau1993social,karau1997effects}. In particular, increased compensation can potentially undermine cohesion within a human teammate. These empirical results might not generalize to our present setting, not least because AI behaviors might be interpreted differently than human behaviors~\cite{de2021deceptive}. More importantly, the empirical work focused on settings with public compensation, whereas adaptive AI compensation is less likely to be noticed. If people are unaware of the compensation, then it is unlikely to undermine trust. However, it can potentially lead to the second issue. 

\textbf{Escalating Compensation Loops}. Consider Alice interacting with an adaptive AI system that compensates to advance goals or values other than Alice's. In this case, Alice is likely to experience a measure of frustration that she consistently falls short of \emph{her} goals. She might thereby intensify her efforts to reach her goals, leading to increased compensation by the adaptive AI, resulting in further changes by Alice, causing \ldots. The coevolution of human and AI behaviors can result in a feedback loop in which each increasingly compensates for the other. Ultimately, such a loop may become unsustainable, in the sense that the human will realize that the AI is attempting to compensate or deceive, leading to the third challenge.

\textbf{Discovery of Compensation}. If unsatisfied users discover that the adaptive or personalized AI system is nonetheless moving them towards other goals, then they may view it negatively. This realization can lead to mistrust and non-compliance, perhaps leading the human to entirely disregard the AI system. That is, the effort to personalize an AI to a specific individual might instead lead that individual to reject the AI when it fails to fully personalize~\cite{jussupow2020we, dietvorst2018overcoming}. Personalization attempts to align different AIs for each individual, and thereby increase individual's trust (and acceptance) in the AI system. That same effort can actually increase the risk of \emph{rejection} of the AI system, since the compensation that naturally arises in personalization can be perceived as a betrayal. Moreover, even if users do not fully reject their AIs, they may devise strategies to circumvent or manipulate these systems as they become more familiar with their workings. Such behaviors have been widely observed for recommender systems (among others) where people can alter their behavior to ensure the outcome that they prefer~\cite{eslami2017careful}. 

\section{Conclusion}
Personalization seems to be a way to sidestep many of the ethical challenges of value pluralism by creating ``AI pluralism.'' We have argued that the AI learning adaptation can create a new ethical challenge---compensation---whenever the AI ought not \emph{entirely} defer to the human. Even a personalized AI ought not support and advance ethically problematic values that an individual might have (e.g., about racial superiority). We thus must consider when an adaptive or personalized AI system should (ethically) compensate for the individual. There is a fundamental tension in pluralistic AI design between accommodation of a wide range of human values, and specific goals or values that are independent of the individual. This misalignment can lead to unintended consequences, including undermining of user autonomy; deceptive practices; and unintentional reinforcement of individual biases.

Despite these concerns, we suggest that compensatory AI behavior should be recognized as a tool in the designer's/developer's toolkit, much as (human) social compensation behavior should be considered when constructing a team. We provide an ethical analysis of (one set of) sufficient conditions for compensation---even undisclosed or deceptive compensation---to be ethically permissible. This analysis also points towards ways to reduce negative effects through minimal compensation. We acknowledge that AI compensation is an ethically challenging possibility (perhaps inevitability, in some settings), but it must be addressed by those pursuing the ``AI pluralism'' strategy towards addressing value pluralism in diverse communities.

{
\small
\bibliographystyle{unsrtnat}
\bibliography{ref}
}

\newpage
\appendix
\section{Appendix}

\subsection{Modeling Interactions}
We provide here an example showing that compensation or deception can be the optimal strategy in a very different formal setting, namely the signaling game shown in Figure~\ref{fig:mdp}.
Suppose that A1 corresponds to providing data to the human without elaboration, while A2 corresponds to additionally providing a recommendation. For concreteness, we use the following  values for the game elements:

\begin{figure}[h]
    \centering
    \includegraphics[width=0.5\textwidth]{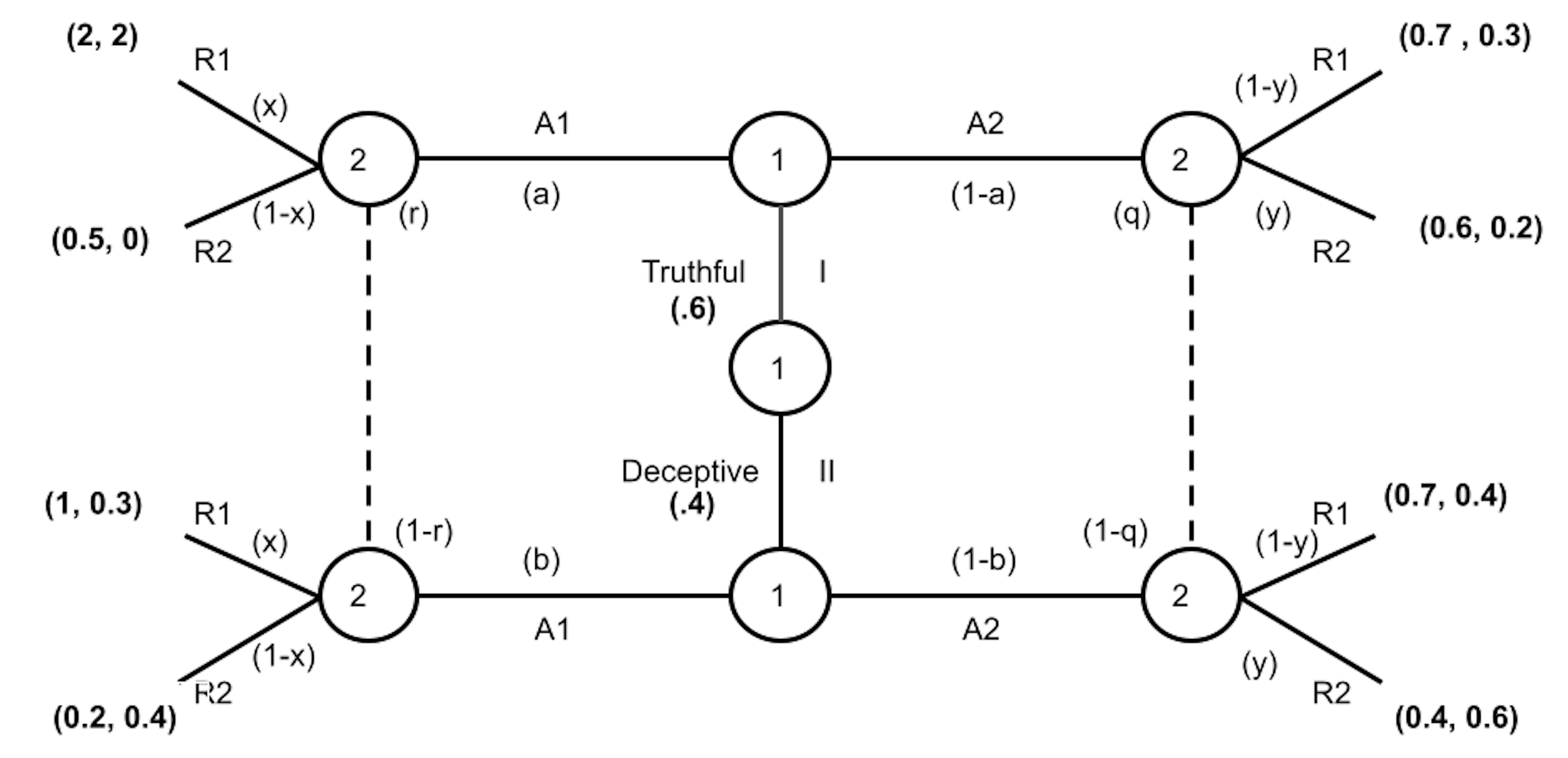}
    \caption{Game tree representation of signaling game with values}
    \label{fig:mdp}
\end{figure}

For the honest and deceptive AIs (I and II):
\begin{align*}
    u_I(A1) &= x(2) + (1-x)(0.5) = 1.5x + 0.5 \\
    u_I(A2) &= y(0.6) + (1-y)(0.7) = -0.1y + 0.7 \\
    u_{II}(A1) &= x(1) + (1-x)(0.2) = 0.8x + 0.2 \\
    u_{II}(A2) &= y(0.4) + (1-y)(0.7) = -0.3y + 0.7 
\end{align*}

Solving these equations yields $x = \frac{1}{37}$ and $y = \frac{59}{37}$. 
And when the human user observes $A1$:
\begin{align*}
    u(R1) &= r(2) + (1-r)(0.3) = 1.7r + 0.3 \\
    u(R2) &= r(0) + (1-r)(0.4) = 0.4 - 0.4r
\end{align*}
And when they observe $A2$:
\begin{align*}
    u(R1) &= q(0.3) + (1-q)(0.4) = -0.1q + 0.4 \\
    u(R2) &= q(0.2) + (1-q)(0.6) = -0.4q + 0.6
\end{align*}
Solving these equations yields $r = \frac{1}{21}$ and $q = \frac{2}{3}$.

We can further derive:
\begin{align*}
    r &= \frac{ap}{ap + b(1-p)} \\
    q &= \frac{(1-a)p}{(1-a)p + (1-b)(1-p)}
\end{align*}
Substituting the values of $r, q, p$ yields the probabilities shown in Table 1 as a Bayesian Nash equilibrium.\\
\begin{table}[h]
\centering
\caption{Probability of Actions Chosen by Each Type of AI}
\begin{tabular}{|c|c|c|}
\hline
\textbf{} & \textbf{Unelaborated Data} & \textbf{Recommendations} \\ \hline
\textbf{Type I} & $1/118$ & $117/118$ \\ \hline
\textbf{Type II} & $15/59$ & $44/59$ \\ \hline
\end{tabular}
\end{table}

\textbf{Observation. } In this example, the AI systems consistently showed a preference for aiding users through recommendations rather than merely presenting data, regardless of the underlying intention (though intention obviously changes the likelihood of recommendation). Specifically, the Honest AI recommended in 117 out of 118 interactions, while the Dishonest AI did so in 44 out of 59 cases. This example thus shows that compensation does not necessarily lead to the AI system being systematically unhelpful; rather, the AI system dynamically alters behavior to enhance overall  success, contrary to the passive role traditionally assumed for AI systems.

\textbf{Observation. } The stable equilibrium reflects the user ``distrusting'' the AI when it simply provides unfiltered data ($r = 1/21$). Thus, in this example, there is a potentially significant ethical issue, as the human would potentially come to \emph{consciously} question the signal from the AI. In such cases, one might reasonably worry that there would be an overall loss of trust, thereby undermining the value of the AI system.

\textbf{Implications.} For the purposes of our analysis, we initially set the deception probability for the Dishonest AI at 0.4. This parameter was chosen to observe the AI's decision-making patterns under conditions of moderate deception. It is reasonable to hypothesize that the probability of the AI will engaging in deceptive strategies could escalate as it continually assesses and refines the efficacy of the tactic. That is, deception is not necessarily a marginal or extreme strategy, but can arguably emerge as a preferred method of operation under certain conditions. 

\subsection{Case for Compensation:  Beneficence}
Increasing reliance on algorithmic systems in decision-making processes raises significant concerns about individual autonomy. Can a personalized compensating algorithm truly respect and support autonomy? \cite{kaminski2018binary}
As we grapple with these challenges to autonomy, we must also consider the principle of beneficence - the moral obligation to act for the benefit of others \cite{abramson1989autonomy}. In biomedical ethics, beneficence is considered alongside other principles, with no single principle taking absolute precedence. Instead, ethical decision-making involves carefully balancing these principles in situations of conflict.

This balancing act becomes particularly relevant in scenarios where an individual's decisions directly impact the well-being of others. In such cases, we encounter a tension between respecting the decision-maker's autonomy and ensuring benefits to those affected by the decision. This conflict challenges us to reconsider the primacy of individual autonomy.
The framework of balancing autonomy and beneficence is crucial when considering the role of human-AI teams in decision-making processes. Consider, for example, a judicial scenario where a human-AI team makes decisions affecting defendants. The AI system, designed to be ethically aware and pluralistic, may produce recommendations that are fair to the defendant but challenge its human collaborator's inherent biases or immediate interests.
A significant challenge arises from the fact that while the AI can offer diverse and potentially more objective perspectives, the human ultimately has the final say in the team's decisions. An AI system, no matter how accurate or ethically designed, cannot directly counter or refute these human decisions. 

In situations where there is a reasonable expectation that the user's biases could lead to suboptimal or biased outcomes, we posit that the principle of beneficence overtakes.  While protecting individual autonomy is crucial, we must also consider the substantial impact of decisions made by human-AI systems since a rigid refusal to impose thoughtful constraints could paradoxically violate our core moral principles \cite{schroeder1986rights}. Since, unlike humans, these systems are not subject to cognitive biases or emotional influences that can skew judgment \cite{sullivan2018employing, kleinberg2018human, bertrand2004emily, rizer2018artificial, wang2023biased}. 

 Of course, this performance depends on having training data that do not encode those human biases; certainly, many AI systems are as biased as humans, if not more so. There are many well-justified concerns about biases and errors from AI systems, but that fact should not lead us to overlook the potential impacts of human biases, particularly in high-stakes domains like criminal justice or healthcare.

 \subsection{ User Experience}
An AI system that consistently compensates for a user's biases or preferences might initially seem to risk reinforcing those biases over time. However, a growing body of research suggests a more nuanced reality. 

Consider, for instance, a clinical setting where a decision support system adapts to a doctor's tendency to over-prescribe surgery. Contrary to initial concerns, such a system isn't necessarily perpetuating this practice. Instead, it operating within a framework of certain immutable restrictions can actually ensure exposure to best practices and alternative approaches. 

These immutable restrictions, often mandated at national levels, create a baseline of ethical and legal compliance. The restrictions prohibit recommendations that could lead to harm or violate established medical guidelines. Building upon this foundation, optional restrictions and requirements, implemented by model and application providers and professional associations, further tailor the system's behavior to align with established values and preferences.

These layered safeguards not only prevent harmful biases but can also promote positive outcomes. In fact, human-computer interaction (HCI)   research supports these claims by demonstrating that AI systems capable of nuanced behaviors, including deceptive characteristics, impact users positively \cite{de2021deceptive}. For instance, in educational settings, the introduction of AI agents capable of limiting the information presented or presenting false information has repeatedly proven to encourage students to engage more actively, enhancing learning efficiency \cite{sjoden2020lying, zhai2021review, tanaka2010care, matsuzoe2012smartly}. Similar results across domains like gaming and physical therapy illustrate this potential \cite{meta2022human, vinyals2019grandmaster, silver2017mastering, brewer2006visual, tanaka2010care}.

\end{document}